# Hybrid Super Intelligence and Polymetric Analysis

V. Dorofeev[1, a)] and P. Trokhimchuck[2, b)]

[1]*Scientific Institute of System Analysis of Russian Academy of Sciences, Moscow, Russia*
[2]*Lesya Ukrainka Volyn National University, 43025, Voly ave., 13, Lutsk, Ukraine*

[a)] *Corresponding author: vladislav.dorofeev@gmail.com*
[b)] *Trokhimchuck.Petro@vnu.edu.ua*

**Abstract.** The problem of possible applications Polymetric Analysis for the resolution problems of artificial Intelligence is discussed. As example the hybrid super intelligence system by N. Moiseev type was selected. The bond between polymetric analysis and hybrid super intelligence system was shown. In operational sense polymetric analysis is more general system. Therefore main principles of Moiseev concept may be unify with the help of polymetric analysis. Main peculiarities of this unification are analyzed.

## INTRODUCTION

The problem of creation universal system of knowledge is connected with problem of complexity of researched information. So, this problem was called the problem of XX-th century by S. Beer [1]. Polymetric analysis (PA) may be represented as variant of resolution this problem [2, 3]. It allows classifying all possible chapters of knowledge, including science, by step of its complexity.

But cybernetics is synthetic science [4, 5]. Therefore, we must been create universal synthetic theory. This theory is Polymetric Analysis.

PA includes the procedure of measurement in finish result of measurement. Basic and derivative measurements are connected with quantitative and qualitative mathematical transformations. First is corresponded the procedure of measurement (arithmetic) the observed quantity, second – the analysis of dimensions (dimensional analysis).

Polymetric analysis is based on idea of triple minimum (particularly scientific, methodical and mathematical). Main principles of PA are criteria of reciprocity and simplicity. The first criterion is the principle of assembling the elements of the corresponding construct into a single system. Second criterion is principle optimality (simplicity complexity) of this assembling.

One of main component of this method, hybrid theory of systems (theory systems with variable hierarchy) show that only ten minimal types system of formalization the knowledge are existed [2, 3].

Moiseev concept of strong artificial intelligence has more deep anthropic nature [6, 9]. This concept is connected with influence human activity on environment. Therefore, in this concept we have three principles: two ecological (Legasov and Efremov) and one system cybernetic (Moiseev) [7]. This concept wss used to the resolution system aspects of COVID-19 [8].

We show that ecological principles may be formulated with help boundary conditions on generalizing mathematical transformations and its combination. Moiseev principle is particularly case of simplicity principle of polymetric analysis.

Thus, Moiseev problem may be represented with help methods of PA too.

# MATERIALS AND METHODS

## Polymetric Analysis as Operational Concept of Computing Science

Polymetric analysis (PA) was created as alternative optimal concept to logical, formal and constructive conceptions of modern mathematics and theory of information [2, 3]. This concept is based on the idea of triple minimum: mathematical, methodological and concrete scientific.

However, one of the main tasks of polymetric analysis is the problem of simplicity-complexity that arises when creating or solving a particular problem or science. It must be open system [3].

In methodological sense, PA is the synthesis of Archimedes thesis: "Give me a fulcrum and I will move the world", and S.Beer idea about what complexity is a problem in cybernetics century, in one system. And as cybernetics is a synthetic science, the problem should be transferred and for all of modern science. Basic elements of this theory and their bonds with other science are represented in FIGURE 1 [3].

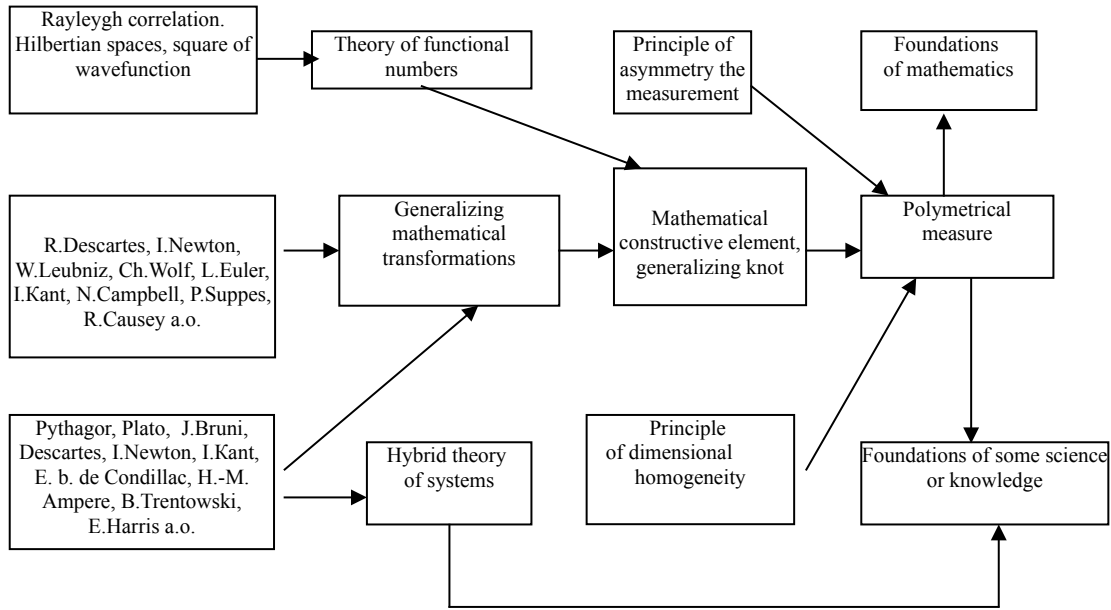

**FIGURE 1.** Schema of polymetric method and its place in modern science [3].

The polymetric analysis may be represented as universal theory of synthesis in Descartian sense. For resolution of this problem we must select basic notions and concepts, which are corresponded to optimal basic three directions of Figure 2. The universal simple value is unit symbol, but this symbol must be connected with calculation. Therefore it must be number. For the compositions of these symbols (numbers) in one system we must use system control and operations (mathematical operations or transformations). After this procedure we received the proper measure, which is corresponding system of knowledge and science.

Therefore the basic axiomatic of the polymetric analysis is was selected in the next form [3]. This form is corresponded to schema of Figure 2.

*Definition 1*. Mathematical construction is called set all possible elements, operations and transformations for resolution corresponding problem. The basic functional elements of this construction are called constructive elements.

*Definition 2*. The mathematical constructive elements $N_{x_{ij}}$ are called the functional parameters:

$$N_{x_{ij}} = x_i \cdot \overline{x}_j, \tag{1}$$

where $x_i$, $\overline{x}_j$ – the straight and opposite parameters, respectively; ▼ – respective mathematical operation.

*Definition 3*. The mathematical constructive elements $N_{\varphi_{ij}}$ are called the functional numbers:

$$N_{\varphi_{ij}} = \varphi_i \circ \overline{\varphi}_j, \tag{2}$$

where $\varphi_i(x_1,...x_n, \overline{x_1}...\overline{x_m},...N_{x_{ij}},...)$, $\overline{\varphi}_j(x_1,...x_n, \overline{x_1}...\overline{x_m},...N_{x_{ij}},...)$ are the straight and opposite functions, respectively; $\circ$ – respective mathematical operation.

*Remark 1.* Functions $\varphi_l$, $\overline{\varphi}_j$ may be have different nature: mathematical, linguistic and other.

*Definition 4.* The mathematical constructive elements $N_{x_{ij}}^d$ are called the diagonal functional parameters:

$$N_{x_{ij}}^d = \delta_{ij} N_{x_{ij}}, \tag{3}$$

where $\delta_{ij}$ is Cronecker symbol.

*Definition 5.* The mathematical constructive elements $N_{\varphi_{ij}}^d$ are called the diagonal functional numbers:

$$N_{\varphi_{ij}}^d = \delta_{ij} N_{\varphi_{ij}}. \tag{4}$$

*Example 1.* If $x_i = x^i$, $\overline{x}_j = x^{-j}$ and $\max\{i\} = \max\{j\} = m$, then $\|\cdot\|$ is diagonal single matrix.

Another example may be the orthogonal eigenfunctions of the Hermitian operator.

*Remark 2.* These two examples illustrate why quantities (1) – (4) are called the parameters and numbers. Practically it is the simple formalization the measurable procedure in its logic and chaos parts [3]. The straight functions correspond the "straight" observation and measurement and opposite functions correspond the "opposite" observation and measurement. This procedure is included in quantum mechanics the Hilbert's spaces and Hermitian operators.

The theory of generalizing mathematical transformations is created for works on functional numbers.

*Definition 6.* Qualitative transformations on functional numbers $N_{\varphi_{ij}}$ (straight $A_i$ and opposite $\overline{A}_j$) are called the next transformations. The straight qualitative transformations are reduced the dimension $N_{\varphi_{ij}}$ on $i$ units for straight parameters, and the opposite qualitative transformations are reduced the dimension $N_{\varphi_{ij}}$ on $j$ units for opposite parameters.

*Definition 7.* Quantitative (calculative) transformations on functional numbers $N_{\varphi_{ij}}$ (straight $O_k$ and opposite $\overline{O}_p$) are called the next transformations. The straight calculative trasformations are reduced $N_{\varphi_{ij}}$ or corresponding mathematical constructive element on $k$ units its measure. The opposite quantitative transformations are increased $N_{\varphi_{ij}}$ or corresponding mathematical constructive element on $l$ units its measure, i.e.

$$O_k \overline{O}_l N_{\varphi_{ij}} = N_{\varphi_{ij}} - k \oplus l. \tag{5}$$

*Definition 8.* Left and right transformations are called transformations which act on left or right part of functional number respectively.

*Definition 9.* The maximal possible number corresponding transformations is called the rang of this transformation:

$$rang(A_i \overline{A}_j N_{ij}) = \max(i, j); \tag{6}$$

$$rang(O_k \overline{O}_p N_{ij}) = \max(k, p). \tag{7}$$

*Remark 3.* The indexes $i,j, k,p$ are called the steps of the corresponding transformations.

Only 15 minimal types of of generalizing mathematical transformations are existed [3].

Basic elements of PA is the generalizing mathematical elements or its various presentations – informative knots. Generalizing mathematical element is the composition of functional numbers (generalizing quadratic forms, including complex numbers and functions) and generalizing mathematical transformations, which are acted on these functional numbers in whole or its elements. Roughly speaking these elements are elements of functional matrixes.

This element ${}_{nmab}^{stqo} M_{ijkp}$ may be represented in next form:

$$_{nmab}^{stqo} M_{ijkp} = A_i \overline{A}_j O_k \overline{O}_p A_s^r \overline{A}_t^r O_q^r \overline{O}_o^r A_n^l \overline{A}_m^l O_a^l \overline{O}_b^l N_{\varphi_{ij}} ,$$
(8)

where $N_{\varphi_{ij}}$ – functional number; $O_k, O_q^r, O_a^l, \overline{O}_p, \overline{O}_o^r, \overline{O}_b^l$; $A_i, A_s^r, A_n^l, \overline{A}_j, \overline{A}_t^r, \overline{A}_m^l$ are quantitative and qualitative transformations, straight and inverse (with tilde), (r) – right and (l) – left.

Polyfunctional matrix, which is constructed on elements (8) is called informative lattice. For this case generalizing mathematical element was called knot of informative lattice [3]. Informative lattice is basic set of theory of informative calculations. This theory was constructed analogously to the analytical mechanics.

Basic elements of this theory are:
1. Informative computability $C$ is number of possible mathematical operations, which are required for the resolution of proper problem.
2. Technical informative computability $C_t = C \Sigma t_i$, where $t_i$ – realization time of proper computation.
3. Generalizing technical informative computability $C_{t0} = k_{ac} C_t$, where $k_{ac}$ – a coefficient of algorithmic complexity.

Basic principle of this theory is the principle of optimal informative calculations : any algebraic, including constructive, informative problem has optimal resolution for minimum informative computability $C$, technical informative computability $C_t$ or generalizing technical informative computability $C_{to}$.

The principle of optimal informative calculations is analogous to action and entropy (second law of thermodynamics) principles in physics.

The principle of optimal informative calculation is more general than negentropic principle the theory of the information [10] and Shennon theorem [11]. This principle is law of the open systems or systems with variable hierarchy. The negenthropic principle and Shennon theorem are the principles of systems with constant hierarchy.

Idea of this principle of optimal informative calculation may be explained on the basis de Broglie formula [12] (equivalence of quantity of ordered and disorder information):

$$S_a / \hbar = S_e / k_B ,$$
(9)

where $S_a$ – action, $\hbar$ – Planck constant, $S_e$ – entropy and $k_B$ – Boltzman constant.

Therefore we can go from dimensional quantities (action and entropy) to undimensional quantity – number of proper quanta or after generalization to number of mathematical operations. Thus, theory of informative calculations may be represented as numerical generalization of classical theory of information.

For classification the computations on informative lattices hybrid theory of systems was created [3]. This theory allow to analyze proper system with point of view of its complexity,

The basic principles of hybrid theory of systems are next: 1) the criterion of reciprocity; 2) the criterion of simplicity.

The criterion of reciprocity is the principle of the creation the corresponding mathematical constructive system (informative lattice). The criterion of simplicity is the principle the optimization of this creation.

The basic axiomatic of hybrid theory of systems is represented below.

*Definition 10.* The set of functional numbers and generalizing transformations together with principles reciprocity and simplicity (informative lattice) is called the hybrid theory of systems (in more narrow sense the criterion of the reciprocity and principle of optimal informative calculations).

***Criterion of the reciprocity*** for corresponding systems is signed the conservation in these systems the next categories:
1) the completeness;
2) the equilibrium;
3) the equality of the number epistemological equivalent known and unknown knotions.

***Criterion of the simplicity*** for corresponding systems is signed the conservation in these systems the next categories:
1) the completeness;
2) the equilibrium;
3) the principle of the optimal calculative transformations.

Criterion of reciprocity is the principle of creation of proper informative lattice. Basic elements of principle reciprocity are various nuances of completeness. Criterion of the simplicity is the principle of the optimality of this creation.

For more full formalization the all famous regions of knowledge and science the parameter of connectedness $\sigma_t$ was introduced. This parameter is meant the number of different bounds the one element of mathematical construction with other elements of this construction. For example, in classic mathematics $\sigma_t = 1$, in linguistics and semiotics $\sigma_t > 1$. The parameter of connectedness is the basic element for synthesis in one system of formalization the all famous regions of knowledge and science. It is one of the basic elements for creation the theory of functional logical automata too.

For classification of systems of calculation hybrid theory of systems was created. This theory is based on two criterions: criterion of reciprocity – principle of creation of proper formal system, and criterion of simplicity – principle of optimality of this creation. For "inner" bond of two elements of informative lattice a parameter of connectedness $\sigma_t$ was introduced. Principle of optimal informative calculation is included in criterion of simplicity.

At help these criteria of reciprocity and simplicity and parameter of connectedness the basic famous parts of knowledge and science may be represent as next 10 types of hybrid systems [3]:

1. The system with conservation all positions the criteria of reciprocity and simplicity for all elements of mathematical construction ($N_{\varphi_{ij}}$ and transformations) is called the *simple system.*
2. The system with conservation the criterion of simplicity only for $N_{\varphi_{ij}}$ is called the *parametric simple system*.
   *Remark* 4. Further in this classification reminder of criteria of reciprocity and simplicity is absented. It mean that these criteria for next types of hybrid systems are true.
3. The system with conservation the criterion of simplicity only for general mathematical transformations is called *functional simple system.*
4. The system with nonconservation the principle of optimal informative calculation and with $\sigma_t = 1$ is called the *semisimple system.*
5. The system with nonconservation the principle of optimal informative calculation only for $N_{\varphi_{ij}}$ and with $\sigma_t = 1$ is called the *parametric semisimple system.*
6. The system with nonconservation the principle of optimal informative calculation only for general mathematical transformations and with $\sigma_t = 1$ is called the *functional semisimple system.*
7. The system with nonconservation the principle of optimal informative calculation and with $\sigma_t \neq 1$ is called *complicated system.*
8. The system with nonconservation the principle of optimal informative calculation only for $N_{\varphi_{ij}}$ is called *parametric complicated system.*
9. The system with nonconservation the principle of optimal informative calculation only for general mathematical transformations and with $\sigma_t \neq 1$ is called *functional complicated system*.
10. The system with nonconservation the criteriums of reciprocity and simplicity and with $\sigma_t \neq 1$ is called *absolute complicated system*.

With taking into account 15 basic types of generalized mathematical transformations we have 150 types of hybrid systems; practically 150 types of the formalization and modeling of knowledge and science.

Only first four types of hybrid systems may be considered as mathematical, last four types are not mathematically. Therefore HTS may be describing all possible system of knowledge. Problem of verbal and nonverbal systems of knowledge is controlled with help of types the mathematical transformations and parameter connectedness [1].

This theory has finite number of types the knowledge formalization systems. And in general, this theory is the theory of open systems with a changeable hierarchy.

Therefore, HTS with its operational nature may be used for all knowledge and culture, including cybernetics and artificial intelligence.

# Architecture of Strong Hybrid Intelligence of N. Moiseev Type

Now we represent the model of strong hybrid intelligence proposed in [7], which is based on N. Moiseev's idea of complex modeling of the noosphere in order to predict the consequences of any external influences on it, including anthropogenic ones [6].

The proposed strong hybrid intelligence architecture is shown in Figure 1 [7].

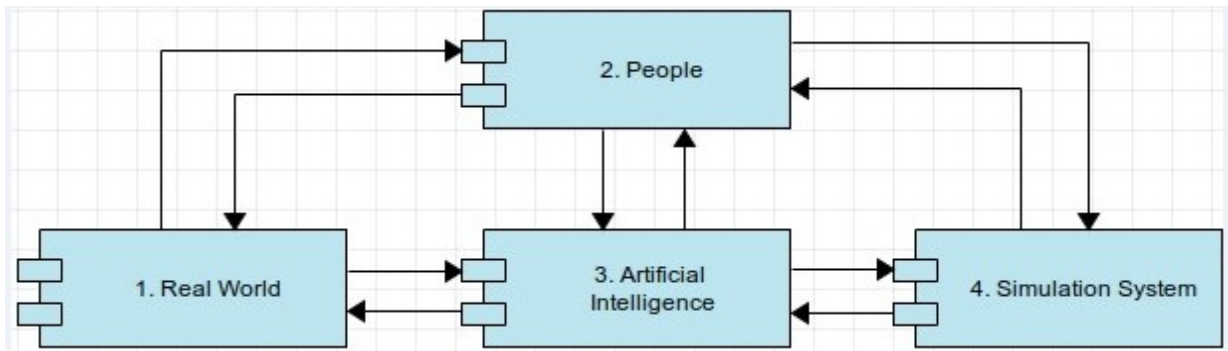

**FIGURE 2**. Architecture the strong hybrid intelligence of N. Moiseev type [7].

1. The real world is information about the problem area, collected using the sensors available to the system. The real world includes both real world objects and automated transactional information processing systems. For example, in the case of a pandemic, contact tracing systems, medical information systems with patient data, etc.
2. People - a team of specialists involved in solving a problem. The team may include subject matter experts, developers, information system operators, etc.
3. Artificial intelligence is an adaptable and developed system for the automated collection and processing of real-world information with an interface for communication with a group of experts, including in natural language.
4. Modeling system - a set of systems for modeling and forecasting the real world with the capabilities of scenario analysis of the consequences of impact on the real world.

The goal of the system of strong hybrid intelligence is the most accurate forecasting of the development of the real world with the possibility of scenario analysis of the consequences of external influences on it.

In [8], this architecture was proposed to solve the COVID-19 problem. A simple experiment was carried out to expand the epidemiological SIRD-model with the parameter "The level of immunity of the population". A similar architecture was used on a larger scale in Canada [7]. As a real world, it used a contact monitoring system and medical information systems with patient data. Artificial intelligence tools were used to refine the parameters of the model system based on real-world data. The modeling system made it possible to carry out scenario analysis of the development of the situation. An international team of specialists ensured the development and use of the system.

In a sense, the proposed architecture can be considered an extension of the concept of "digital twins", developed in the world since 2002, to include the human factor. And at present, the main successes in the field of solving complex problems are associated with precisely such systems [6].

The architecture of the proposed system itself assumes its transparency and controllability, since the consequences of the supposed impacts on the real world are checked on the modeling system. But taking into account the complexity of the real world and the limited possibilities of methods for its modeling, there are no complete guarantees of the safety of such a system. The following principles can be used to improve security.

1. Legasov's principle: a technostructure uncontrolled by society threatens the global security of mankind.
   The first principle concerns the problem of control of the technostructure, which controls the system of strong artificial intelligence, by the society. It appeared on the basis of V. Legasov's notes on the circumstances of the Chernobyl accident [7]. In relation to the proposed system of strong hybrid intelligence, it means the following: transparency and availability of the information systems used.
2. Efremov's principle: anthropogenic changes in the environment, the rate of which exceeds the physical, biological and social mechanisms of adaptation to them, carry the risks of destroying life.

The second principle is related to the ability of living systems to adapt to changes in the external environment, which the hybrid super intelligence system plans to carry out. If the speed of adaptation is not enough, then it is better to refrain from making changes. It was formulated on the basis of an episode with a visit by earthlings to the planet Zirda from the book "The Andromeda Nebula" by I. Efremov [7]. Only poppies remained on the planet due to the underestimation of the danger by its inhabitants of the use of low-power ionizing radiation, contributing to a high rate of uncontrolled mutations.

3. Moiseev's principle: the complexity of the models of the world should be comparable to the complexity of the problems they are designed to solve.

The third principle is designed to prevent the use of primitive models to solve complex problems. It was formulated on the basis of the works of N. Moiseev [6]. Despite some advances, current modeling capabilities of real-world systems are limited in their ability to accurately reproduce the real world. And often researchers forget about this and go far from the real world in their theoretical constructions.

## Hybrid Super Intelligence of N. Moiseev Type and Polymetrical Analysis

We can analyze PA and N. Moiseev system with point of conditions, which are formulated for the general theories (theories of everything) [2, 3]:
1. It must be open theory or theory with variable hierarchy.
2. This theory must be having minimal number of principles.
3. It must based on nature of mathematics (analysis, synthesis and formalization all possible knowledge).
4. We must create sign structure, which unite verbal and nonverbal knowledge (mathematical and other) in one system.
5. We must have system, which is expert system of existing system of knowledge and may be use for the creation new systems of knowledge.
6. Principle of continuity must be true for all science.

These conditions must be used for the creation any dynamic science, which can be presented as open system.

Polymetric analysis fully satisfies these conditions, the system of N. Moiseev – partially.

Polymetrical Analysis is more general system as cybernetics and strong hybrid intelligence of N. Moiseev type. It is operational system, which is included the procedure of measurement with help generalizing mathematical transformations. Generalized constructive element (8) is term of polyfunctional matrix. But computer processors are using matrix calculation [13, 14]s. Therefore PA may be represented as functional expansion of computer processor, which are include the procedures of collection and processing of information using generalized mathematical transformations and criteria of reciprocity and simplicity.

Strong hybrid intelligence of N. Moiseev type has more narrow scope. Therefore it is more anthropic system as PA [6, 9]. Roughly speaking Legasov, Efremov principles are anthropic principles and only Moiseev principle is connected with procedure of system formalization.

Legasov principle is ecological principle and has more long history.These are the problems of chemical, nuclear and atmospheric pollution of the environment. These problems were raised by physicists themselves such as A. Einstein, N. Bohr, A. Sakharov, philosophers– B. Russell and other (Russel-Einstein manifesto, Pugwash Conferences on Science and World Affairs and Greenpeece) [15]. The same political movement of the greens arose. Based on this, the international independent non-governmental environmental organization Greenpeace was established in 1971 in Canada [15].

The organization's focus is on issues such as global climate change, deforestation from the tropics to the Arctic to the Antarctic, overfishing, commercial whaling, radiation hazards, renewable energy and resource conservation, hazardous chemical pollution, sustainable agriculture, economy, preservation of the Arctic.

Therefore Legasov and Efremov principles have ecological nature. Legasov principle is characterized the irreversible change in nature, Efremov principle is corresponded to the establishing the boundaries of this irreversibility.

From mathematical point of view the Legasov principle may be represented as boundary conditions on corresponding qualitative transformations, Efremov principle imposes an additional cyclic condition, which must be imposed on the corresponding mathematical construct. For the Efremov principle, a reversible reproducible feedback is important, which can be set both through inverse mathematical transformations and through their combinations.

The division of complexity into subject and mathematical is quite conventional. In polymetric analysis, it has a systematic form with an emphasis on calculation. The fact that the complexity of information processing needs to be carried out through calculations was pointed out by Kasti [16].

The Moiseev principle, the principle of correspondence between the complexity of the formalized (modeled) area and formalization (modeling) methods may be represented as equivalence three provisions of the principle of reciprocity, especially the first two provisions with the third. Really in computer science problem of complexity must reduced to problem of complexity the calculation in more general sense. For PA it must be more optimal principle with calculation point of view.

## CONCLUSIONS

1. Basic concepts of Polymetric Analysis as universal system of formalization the knowledge are analyzed.
2. We show that hybrid theory of systems as finite system of formalization the knowledge may be used for the classification knowledge by step of its simplicity-complexity.
3. Strong hybrid intelligence of Moiseev type is analyzed.
4. Operational bond of Polymetric Analysis and N. Moiseev system is discussed on the basis conditions, which was formulated for the creation universal theories (theories of everything).

## ACKNOWLEDGMENTS

We would like to express our gratitude to N. Nepeyvoda, A Ershov, N. Moiseev, A. Kukhtenko, A. Ivakhnenko, W. L. Dunin-Barkowski, V. Skorobagatko, S. Kuzmich, F. T. Grier, E. E. Harris, M. Bunge A. Kifishin, S. Illarionov, A. Cherezov, N. Shkurina, L. Akimova, A. Pavlenko, I. Stoyanova, V. Bonch-Bruevich, D. Zubarev, Yu. Klimontovich, A. Sakharov, I. Shafarevich, S. Nikolskiy, N. Gangan, N. Fedorenko, V.Obolensky, L. Morozov, Yu.Tulyakov, P. Kazakov , N. Kolpakov, I. Kanterov, R. Yurkin, V. Minin, V. Samarsky. We also express gratitude to  participants of N. Nepeyvoda and A. Ershov computer seminars; Yu. Samoylenko cybernetical seminars; W. Dunin-Barkowski neuroscience and artificial intelligence seminars; V. Ginzburg on theoretical physics seminars in P. Lebedev physical institute and "round tables" in the chamber of the Leninian library (now National Russian Library in Moscow) for the information provided to us, as well as for the discussion on a number of issues that are presented in the proposed paper.

The work financially supported by State Program of SRISA RAS No. 0065-2019-0003 (AAA-A19-119011590090-2).## REFERENCES

1. S. Beer, "We and complexity of modern world," in *Cybernetics today: problems and propositions* (Znaniye, Moscow, 1976), pp. 7–32.
2. P. P. Trokhimchuck, "S. Beer centurial problem in cybernetics and methods of its resolution," in *Advanced in computer science 7(5)*, edited by M. Singla (AkiNik Publications, New Delhi, 2020), pp. 87–118.
3. P. P. Trokhimchuck, *Theories of Everythings: Past, Present, Future* (Lambert Academic Publishing, Saarbrukken, 2021), pp. 260.
4. F. H. George, *Philosophical Foundations of Cybernetics* (Abacus Press, London, 1976), pp. 157.
5. A. I. Kukhtenko, *Cybernetics and Fundamental Science* (Naukova Dumka, Kiev, 1987), pp. 144.
6. N. N. Moiseev, *Development algorithms* (Nauka, Moscow, 1987), pp. 302.
7. V. P. Dorofeev, "Hybrid Super Intelligence and Principles of Legasov, Efremov and Moiseev," in *Actual Problems of Fundamental science-2021,* Conference Proceedings IV, edited by P. P. Trokhimchuck *et al.* (Vezha-Print, Lutsk, 2021), pp. 47–50.
8. V. Dorofeev, A. Lebedev, V. Shakirov and W. Dunin-Barkowski, "Super Intelligence to solve COVID-19 Problem," in *Advances in Neural Computation, Machine Learning and Cognitive Research IV*, pp. 293–300 (2020).
9. J. D. Barrow and F. Tippler, *The Anthropic Cosmological Principle* (University Press, Oxford, 1986), pp. 676.
10. L. Brillouin, *Science and Information Theory* (Courier Corporation, New York, 2004), pp. 392.
11. C. E. Shannon, The Bell System Technical Journal **27**, pp. 379–423, pp. 623–656 (1948).

12. L. de Broglie, "Thermodynamics of isolated point (Hidden thermodynamics of particles)," in *L. de Broglie. Collected papers 4* (Print-Atel'e, Moscow, 2014), pp. 8–111.
13. G. N. Nillson, *The Quest for Artificial Intelligence: A History of Ideas and Achievements* (Cambridge University Press, New York, 2010), pp. 562.
14. V. S. Emelynov, "Pugwash movement," in *About science and civilizations. Memories, thoughts and reflections of a scientists,* edited by I. G. Usachev (Mysl, Moscow, 1986), pp. 32–38.
15. N. Nepeyvoda, *Programming Styles and Techniques* (Internet University of Information Technologies, Moscow, 2005), pp. 320.
16. J. Kasti, *Big systems. Connectivity, complexity and catastrophe* (Wiley @ Sons, New York, 1979), pp. 216.